\documentclass[letterpaper, 10 pt, conference]{ieeeconf}  % Comment this line out if you need a4paper

\IEEEoverridecommandlockouts                              
\usepackage{amsmath} % assumes amsmath package installed
\usepackage{algorithm}
\usepackage{algpseudocode}
\usepackage{subcaption}
\usepackage{hyperref}
\usepackage{graphicx}
\usepackage{listings}
\lstdefinelanguage{PDDL}{
    keywords={and, not, action, param, pre, eff, stream, inp, dom, out, cert, Initial, literal, Goal, Baseline, -, Co, Model, MCQA, with},   % General keywords
    keywordstyle=\bfseries,                         % Bold keywords
    sensitive=true,
    morecomment=[l]{;},                             % Comments start with ;
    commentstyle=\color{gray}\itshape
}

\newcommand{\methodname}{\textsc{CoCo-TAMP}}

\newcommand{\eg}{\textit{e.g.}}
\newcommand{\ie}{\textit{i.e.}}

\title{\LARGE \bf
Large-Language-Model-Guided State Estimation for Partially Observable Task and Motion Planning}

\author{Yoonwoo Kim$^{1}$, Raghav Arora$^{1}$, Roberto Martín-Martín$^{1}$, Peter Stone$^{1}$,\\ Ben Abbatematteo$^{1 \dag}$, Yoonchang Sung$^{2 \dag}$% <-this % stops a space
\thanks{$^{1}$The University of Texas at Austin }%
\thanks{$^{2}$Nanyang Technological University}%
\thanks{$^{\dag}$ Equal advising}% <-this % stops a space
\thanks{Correspondence: yoonchang.sung@ntu.edu.sg}
}

\begin{document}

\maketitle
\thispagestyle{empty}
\pagestyle{empty}

%%%%%%%%%%%%%%%%%%%%%%%%%%%%%%%%%%%%%%%%%%%%%%%%%%%%%%%%%%%%%%%%%%%%%%%%%%%%%%%%
\begin{abstract}

Robot planning in partially observable environments, where not all objects are known or visible, is a challenging problem, as it requires reasoning under uncertainty through partially observable Markov decision processes. During the execution of a computed plan, a robot may unexpectedly observe task-irrelevant objects, which are typically ignored by naive planners. In this work, we propose incorporating two types of common-sense knowledge: (1) certain objects are more likely to be found in specific locations; and (2) similar objects are likely to be co-located, while dissimilar objects are less likely to be found together. Manually engineering such knowledge is complex, so we explore leveraging the powerful common-sense reasoning capabilities of large language models (LLMs). Our planning and execution framework, CoCo-TAMP, introduces a hierarchical state estimation that uses LLM-guided information to shape the belief over task-relevant objects, enabling efficient solutions to long-horizon task and motion planning problems. In experiments, CoCo-TAMP achieves an average reduction of $62.7\%$ in planning and execution time in simulation, and $72.6\%$ in real-world demonstrations, compared to a baseline that does not incorporate either type of common-sense knowledge.

\end{abstract}

%%%%%%%%%%%%%%%%%%%%%%%%%%%%%%%%%%%%%%%%%%%%%%%%%%%%%%%%%%%%%%%%%%%%%%%%%%%%%%%%
\section{Introduction}

% generator - verifier add 
Robots performing long-horizon manipulation tasks must reason over discrete decisions, such as which objects to interact with, and continuous motions for manipulation and navigation. Task and motion planning (TAMP)~\cite{garrett2021integrated} provides a principled approach for such problems. However, in realistic settings with uncertainty over object poses and occlusions, plans from deterministic TAMP solvers can fail. This work addresses the partially observable task and motion planning (PO-TAMP) problem, aiming to enable robots to effectively plan to manipulate objects that may not be directly visible due to partial observability.

We introduce \methodname{}, a TAMP system that uses large language models (LLMs) to provide common-sense priors and co-location cues that shape beliefs during planning and execution. While LLMs have limitations as planners~\cite{liu2023llm+, valmeekam2023planning}, they can serve as a rich, approximate source of knowledge ~\cite{guan2024task, kambhampati2024position}. LLM-Modulo~\cite{kambhampati2024position} provides a framework that leverages the strength of LLMs with external model-based verifiers in a ``generate and verify" loop, allowing flexibility in problem specification while guaranteeing the completeness of the framework. Following this paradigm, \methodname{} queries an LLM to form priors over rooms/surfaces via multiple-choice question answering (MCQA) and uses LLM sentence embeddings to build a similarity-based co-location model that propagates evidence across objects. Beliefs are maintained in a hierarchical Bayesian filter over rooms, surfaces, and continuous poses, with a visibility-aware observation model that addresses misses under partial coverage and a lightweight “co-location toggler” that disables co-location when semantics suggest broad dispersion (\eg, switches). Integrated with a belief-space planner, these components yield efficient information gathering and execution. In simulation and on a real robot, \methodname{} reduces cumulative planning and execution time by \(\mathbf{62.7\%}\) and \(\mathbf{72.6\%}\), respectively, compared to a baseline without LLM priors or co-locations.

In summary, this paper makes the following contributions:
\begin{itemize}
  \item We propose an interleaved planning–execution framework for PO-TAMP that leverages LLMs to provide commonsense knowledge. By providing informative priors and guiding belief updates, \methodname{} enables more accurate object state estimation and makes belief-space planning practical under partial observability.

  \item We demonstrate the effectiveness of our approach through large-scale household simulations and real-world robot experiments. Across diverse environments, our framework substantially reduces planning and execution time compared to baselines, and maintains robustness even in adversarial settings where commonsense priors are misleading.
\end{itemize}

\section{Problem Description}\label{sec:prob}
We assume a known semantic layout (rooms and surfaces) obtained from semantic SLAM system~\cite{hughes2022hydra}, but only partial information about objects is available. The environment contains $O$ manipulable objects, $R$ rooms, and $S$ surfaces. For each task-relevant object $k\in\{1,\dots,K\}$, let $x_{r,t}^k\in\mathcal{R}$, $x_{s,t}^k\in\mathcal{S}$, and $x_{p,t}^k\in SE(3)$ denote its room, surface, and pose at time $t$, with beliefs $\text{bel}(x_{r,t}^k)$, $\text{bel}(x_{s,t}^k)$ be categorical and $\text{bel}(x_{p,t}^k)$ be continuous. We are given initial beliefs $\text{bel}(x_{r,0}^k)$, $\text{bel}(x_{s,0}^k)$, and $\text{bel}(x_{p,0}^k)$ over $K<O$ task‑relevant objects.

Building on SS-Replan~\cite{garrett2020online}, we model PO‑TAMP as a hybrid discrete–continuous belief‑space stochastic shortest‑path problem~\cite{bertsekas1991analysis} where actions have strictly positive cost and the minimum‑cost plan maximizes success likelihood. \methodname{} takes a TAMP specification $(\mathcal{O},\mathcal{P},\mathcal{I},\mathcal{G},\mathcal{A})$, where $\mathcal{O}$ is the set of manipulable objects, $\mathcal{P}$ the predicates, $\mathcal{I}$ the initial literals, $\mathcal{G}$ the goal literals, and $\mathcal{A}$ a finite set of parameterized actions with preconditions and effects. Throughout planning and execution, \methodname{} maintain beliefs over $(x_{r,t}^k,x_{s,t}^k,x_{p,t}^k)$ for each task‑relevant object $k$ and triggers replanning whenever execution failures occur.

PDDLStream~\cite{garrett2020pddlstream} was used as the underlying TAMP planner to couple symbolic actions with streams that sample and certify continuous variables (\eg, poses and trajectories). In addition to standard pick, place, and move actions, we include an observation action \texttt{detect}. The \texttt{detect} action is parameterized by the object \texttt{?o}, surface \texttt{?s}, room \texttt{?r}, sampled object pose from $\text{bel}(x_{p,0}^k)$ \texttt{?pb}, base configuration \texttt{?bq}, head configuration \texttt{?hq}, and head trajectory \texttt{?ht} of the robot. Following the self-loop MDP heuristic in SS-Replan, the cost of taking the \texttt{detect} action (\texttt{DetectCost}) is set inversely to the belief over the object's state, which serves as a proxy for detection likelihood (\ie, cheaper when belief mass and visibility predict success). Concretely, this cost is computed as $\frac{c_a}{\text{bel}(x_{r,t}^k, x_{s,t}^k, x_{p,0}^k)}$ where $c_a = 1$ and the resulting cost steers the planner toward informative views. 

We are particularly interested in improving the efficiency of solving PO-TAMP problems by achieving accurate beliefs guided by LLM-driven initializations and object co-location models. To this end, we evaluate \methodname{} using two metrics: the cumulative planning and execution time until successful task completion, and the number of replanning iterations required to solve the problem. The latter metric is motivated by the fact that the former can vary significantly depending on implementation details, system specifications, and the task itself. As such, the number of execution failures serves as a useful proxy metric for evaluating efficiency. 

\section{COCO-TAMP}
\begin{figure*}[t]
\centering
\includegraphics[width=0.8\textwidth]{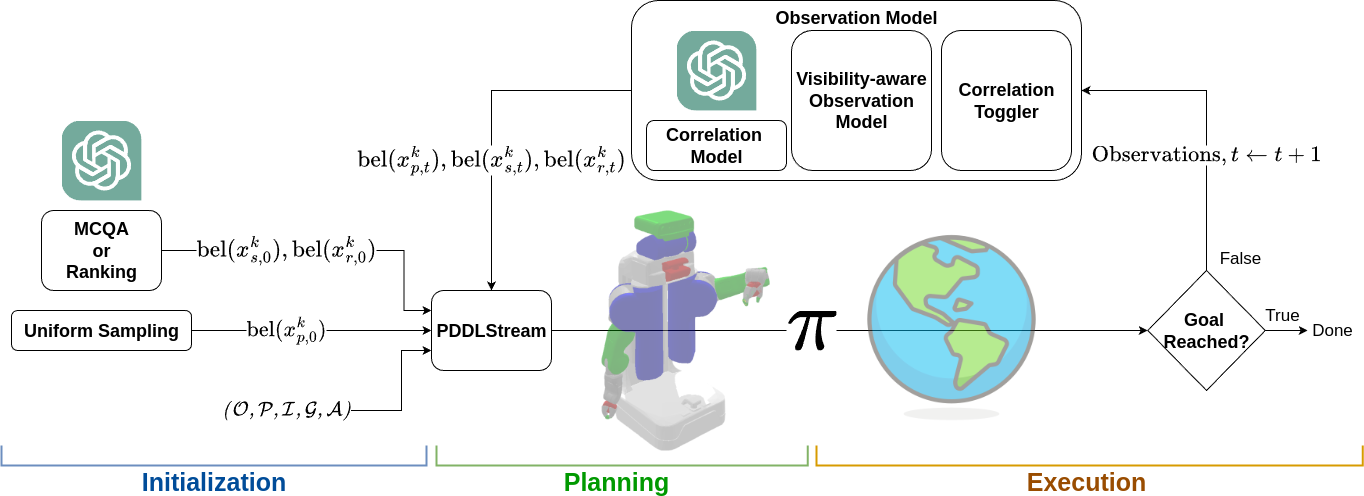}
\caption{The initial beliefs about the semantic locations of objects, $\text{bel}(x_{r,0}^k)$ and $\text{bel}(x_{s,0}^k)$, are derived from LLMs, while the initial beliefs about their poses, $\text{bel}(x_{p,0}^k)$, are uniformly distributed across all surfaces. The TAMP problem specification $(\mathcal{O, P, I, G, A})$, where the beliefs are incorporated into the cost of the observation action, is provided to a TAMP planner PDDLStream. The planner outputs a plan, which the robot executes. Upon executing the observation action, the co-location toggler determines whether to use the co-location model or not, based on the observed object. Then, the beliefs are updated using the proposed observation model, and the planner replans with the updated beliefs. Planning and execution are repeated until the goal state $\mathcal{G}$ is reached. Further implementation details, derivations, additional experimental results, and link to the code are available at our project page: \href{https://coco-tamp.github.io}{\texttt{https://coco-tamp.github.io}}.}\label{fig:1}
\end{figure*}

The overview of \methodname{} is presented in Figure~\ref{fig:1}, which illustrates an integrated planning and execution framework. LLM-guided initial belief generation is discussed in Section~\ref{sub:initial_belief}. Section~\ref{sub:state_est} describes how the beliefs are updated during execution using a hierarchical Bayes filter with LLM-informed semantic similarity between objects.

\subsection{Initial belief generation with LLMs}\label{sub:initial_belief}
LLMs trained on internet-scale data possess common-sense knowledge about typical object placements. Leveraging this capability, \methodname{} utilizes LLMs to generate prior beliefs over rooms and surfaces, denoted by $\text{bel}(x_{r,0}^k)$ and $\text{bel}(x_{s,0}^k)$.

Prior work~\cite{kumar2023conformal,vishwakarma2024improving,su2024api} has employed MCQA for uncertainty quantification in LLMs. We adopt their approach by formulating the most likely room and surface selection problem as an MCQA task. Possible rooms and surfaces where the object could be located are paired with single-character options (\ie, A, B, C, ...). The LLM is asked to choose the character corresponding to the room and surface where the object is most likely to be located. An example input prompt given to the LLM to estimate $\text{bel} (x_{r,0}^k)$ is presented below, and the corresponding prompt to estimate $\text{bel}(x_{s,0}^k)$ omitted, as it follows an analogous structure.\\
\texttt{Predict the location of a toaster.\\
(A) Kitchen\\
(B) Bathroom\\
(C) Livingroom\\
(D) Garage\\
Return the letter that represents the location:}

The probabilities over potential answers can be derived from the LLM’s next-token prediction mechanism by evaluating the log probabilities of each possible answer’s tokens when appended to the prompt. Consequently, given a one-shot input prompt similar to the one above, we obtain logit scores corresponding to the option choices (\ie, A, B, C, $\dots$). These logit scores are then passed through a softmax function, which converts them into valid probabilities.

\subsection{Object state estimation}\label{sub:state_est}
To obtain the probability of successful detection, we jointly estimate the semantic location (\ie, surface, room) and the pose of the objects while accounting for the \emph{visibility} of the location. Formally, we aim to compute the posterior probability of the $k$-th object's semantic location and pose, denoted as $\text{bel}(x_{r,t}^k, x_{s,t}^k, x_{p,t}^k)$, using the recursive Bayesian filtering framework.

Objects are on surfaces, and surfaces are in rooms. Consequently, the belief distribution $\text{bel}(x_{r,t}^k, x_{s,t}^k, x_{p,t}^k)$ is factored into three conditional terms: 
\begin{itemize}
    \item $\text{bel}(x_{p,t}^k)=P(x_{p,t}^k|x_{s,t}^k, x_{r,t}^k, z_{p,t}^k)$
    \item $\text{bel}(x_{s,t}^k)=P(x_{s,t}^k|x_{r,t}^k, \mathcal{Z}_{s,t}, v_{s,t})$
    \item $\text{bel}(x_{r,t}^k)=P(x_{r,t}^k|\mathcal{Z}_{r,t}, v_{r,t})$
\end{itemize}

The observation models include both continuous and categorical components. Specifically, $z_{p,t}^k$ denotes the continuous observation of the $k$-th object’s pose, while $\mathcal{Z}_{s,t}$ and $\mathcal{Z}_{r,t}$ correspond to categorical observations associated with surfaces and rooms, respectively. The associated coverage terms, $v_{r,t}, v_{s,t} \in [0, 1]$, quantify the visibility of surface- and room-level observations. Further details are provided in a later section.

$P(x_{p,t}^k|x_{s,t}^k, x_{r,t}^k, z_{p,t}^k)$ denotes the probability that the $k$-th object is at a particular pose at time $t$, given that it is located on a specific surface, in a particular room, and the pose observation.
$P(x_{s,t}^k|x_{r,t}^k, \mathcal{Z}_{s,t}, v_{s,t})$ represents the probability that the $k$-th object is located on a particular surface, given that it resides in a specific room and the corresponding surface observations.
Lastly, $P(x_{r,t}^k|\mathcal{Z}_{r,t}, v_{r,t})$ represents the probability of the $k$-th object being in a particular room, given room observations.
In this work, the first term is estimated using a particle filter, and the latter two are estimated using discrete Bayes filters. 

Due to the similarity in their derivations, we present the full equations related to room-level belief updates $\text{bel}(x_{r,t}^k)$ in the main text, and omit the analogous equations for surface-level beliefs $\text{bel}(x_{s,t}^k)$.

\subsubsection{Estimating object's semantic location}
For the $k$-th object, we apply hierarchical discrete Bayes filters to estimate its surface and room locations, respectively. Since we consider only stationary objects, typical motion models in recursive Bayesian filtering are unnecessary. For notational convenience, we use $\overline{\text{bel}}$ to denote the predicted belief in the following Bayes filtering equations, which is essentially the same as the prior belief at time $t-1$. The recursive Bayes filter equation for room category estimation is as follows:
\begin{align}
    \text{bel}(x_{r,t}^k) &= P(x_{r,t}^k|\mathcal{Z}_{r,t}, v_{r,t}),\\
    &= \eta P(\mathcal{Z}_{r,t}|x_{r,t}^k, v_{r,t}) \overline{\text{bel}}(x_{r,t}^k),
\end{align}
where $\eta$ represents a normalization constant. 

We denote $v_{r,t}, v_{s,t} \in [0, 1]$ as the visibility of a room and a surface, respectively, from the robot’s field of view. These visibility values are computed using the particles sampled to estimate the posterior distribution $P(x_{p,t}^k|x_{s,t}^k, x_{r,t}^k, z_{p,t}^k)$. Let $\mathcal{S}$ be the set of surfaces in a room. For each surface $s \in \mathcal{S}$, let $n_{s,0}$ be the number of particles initially drawn uniformly on $s$, and let $n_{s,t}^{\text{(seen)}}$ be the number of particles detected on $s$ at time $t$ from robot's field of view when applying the \texttt{detect} action. Then, $v_{s,t}$ and $v_{r,t}$ are calculated as:
\begin{align}
    v_{s,t} = \frac{n_{s,t}^{\text{(seen)}}}{n_{s,0}}, v_{r,t} = \frac{\sum_{s\in \mathcal{S}} n_{s,t}^{\text{(seen)}}}{\sum_{s \in \mathcal{S}} n_{s,0}}.
\end{align}

The visibility is incorporated into our observation model to account for partial observability, where occlusions or limited field of view may prevent successful observations. Consequently, when reasoning about an object's semantic location (\eg, a specific surface within a room), a failed observation does not definitively indicate the object's absence.

Following COS-POMDP~\cite{zheng2022towards}, the categorical observation set for a set of objects over semantic locations can be factored into a product of individual observations for each object using the conditional independence assumption~\cite{dawid1979conditional}, effectively separating the visibility-aware observation model for the object of interest during the process. 

The conditional independence assumption suggests that the observation of an object $z_t^i$ at the current time $t$ is conditionally independent of all other objects' observation, given the object's state $x_t^i$. For example, given two objects $i, j \in K$ where $i \neq j$,
\begin{align}
    P(z_t^i,z_t^j|x_t^j) = P(z_t^i|x_t^j)P(z_t^j|x_t^j).  
\end{align}

Essentially, this assumption implies that the observation of object $i$ does not provide additional information about the observation of object $j$ once the state of object $j$ is already known. Under this assumption, the observation model for updating $\text{bel}(x_{r,t}^k)$ can be written as:
\begin{equation}
\begin{aligned}
    P(\mathcal{Z}_{r,t}|x_{r,t}^k, v_{r,t}) = P(z_{r,t}^k|x_{r,t}^k, v_{r,t}) \\ \times \prod_{j=1, j \neq k}^{K} P(z_{r,t}^j|x_{r,t}^k).
\end{aligned}  
\end{equation}
We refer to $P(z_{r,t}^k|x_{r,t}^k, v_{r,t})$ as \emph{visibility-aware observation models}. The visibility-aware observation model is defined as follows:
\begin{itemize}
    \item \textbf{Case 1:} Observation of the location where the object is placed (\ie, $z_{r,t}^k = x_{r,t}^k$).
\end{itemize}
\begin{align}
    P(z_{r,t}^k \mid x_{r,t}^k, v_{r,t}) =
    \begin{cases}
        (1 - p_{\mathrm{fn}})\,v_{r,t}, & \text{Det.},\\
        (1 - v_{r,t}) + v_{r,t}\,p_{\mathrm{fn}}, & \text{No Det.}.
    \end{cases}
\end{align}
\begin{itemize}
    \item \textbf{Case 2:} Observation of the location where the object is \emph{not} placed (\ie, $z_{r,t}^k \neq x_{r,t}^k$).
\end{itemize}
\begin{align}
    P(z_{r,t}^k \mid x_{r,t}^k, v_{r,t}) =
    \begin{cases}
        p_{\mathrm{fp}}\,v_{r,t}, & \text{Det.},\\
        1 - v_{r,t} p_{\mathrm{fp}}, & \text{No Det}.
    \end{cases}
\end{align}

The false negative and false positive are denoted as $p_{\text{fn}}$ and $p_{\text{fp}}$, respectively, which are set to $0.01$ in our experiments.
\subsubsection{co-location model}\label{subsub:co-model}
The visibility-aware observation model itself does not provide any information about the semantic location of another object. However, in real-world environments, similar objects are often stored together, while dissimilar objects are typically stored apart. To address this gap, we propose a co-location model that leverages LLM embeddings. 

Let $\mathcal{R}$ be a set of rooms in an environment. Using the law of total probability, $P(z_{r,t}^j|x_{r,t}^k)$ can be rewritten as:
\begin{align}
    P(z_{r,t}^j|x_{r,t}^k) = \sum_{x_{r,t}^j \in \mathcal{R}} P(z_{r,t}^j|x_{r,t}^j) P(x_{r,t}^j | x_{r,t}^k).
\end{align}
We refer to $P(x_{r,t}^j | x_{r,t}^k)$ as \emph{co-location models}, which incorporate the similarity between objects $j$ and $k$.

We compute the similarity between objects by applying cosine similarity to their respective LLM embeddings. Specifically, for each object, we use an LLM to generate three sentences describing its common uses, concatenate these sentences into a single textual prompt, and then obtain an embedding of that prompt from an LLM embedding model. See Appendix~\ref{appx:co_model_prompt} for a prompting example. The cosine similarity between the resulting embeddings serves as the similarity score $\texttt{sim}(j,k) \in [-1,1]$.

The co-location models  exhibit the following properties:
\begin{itemize}
    \item If objects $j, k$ are identical ($\texttt{sim}(j,k) = 1$), the co-location models reduce to the Kronecker delta.
    \item If objects $j, k$ have no co-location  ($\texttt{sim}(j,k)=0$), the co-location models become a uniform distribution.
    \item If objects $j, k$ are very different  ($\texttt{sim}(j,k)=-1$), the co-location models are the complement of the Kronecker delta.
\end{itemize}

If $\texttt{sim}(j,k) \geq 0$, the co-location models interpolate between the uniform distribution($u(a, b)$) and the Kronecker delta ($\delta(a,b)$), depending on the similarity between the two objects. If $\texttt{sim}(j,k) \leq 0$, the co-location models interpolate between the normalized complement of the Kronecker delta and the uniform distribution, based on the similarity of the two objects.
Formally, we define the co-location model $ P(x_{r,t}^j|x_{r,t}^k)$ for $R$ rooms in $\mathcal{R}$ as shown below.

Let $\delta_{jk} = \delta(x_{r,t}^j, x_{r,t}^k)$, $\bar\delta_{jk} = \frac{1-\delta(x_{r,t}^j, x_{r,t}^k)}{R-1}$, and $u_{jk} = u(x_{r,t}^j, x_{r,t}^k)$. Then
\begin{align}
P(x_{r,t}^j \mid x_{r,t}^k) &=
\begin{cases}
\texttt{sim}(j,k)\,\delta_{jk} \\+ (1-\texttt{sim}(j,k))\,u_{jk}, \texttt{ sim}(j,k)\!\geq\!0, \\
\texttt{abs}(\texttt{sim}(j,k)) \bar\delta_{jk} \\+ (1+\texttt{sim}(j,k))\,u_{jk}, \texttt{ sim}(j,k)\!\leq\!0.
\end{cases}
\end{align}

As a result, observing an object increases the belief of similar objects sharing that semantic location, and vice versa for dissimilar objects.

Some objects may tend to be distributed across different areas of the environment; for instance, observing a light switch in a bedroom does not imply that all light switches are located in the bedroom. To accommodate such variability, \methodname{} leverages the LLM to decide whether to enable the co-location model based on the semantics of the observed object. An example prompt used for this decision process is provided in Appendix~\ref{appx:toggle_prompt}.

\subsubsection{Estimating object pose}\label{subsub:obj_pose}
The final step in object state estimation is to compute the belief about the object's pose, $\text{bel}(x_{p,0}^k)$. To update $\text{bel}(x_{p,0}^k)$, two types of continuous observation models are introduced: one for when the object is visible, and the other for when the object is not visible. These observation models are applied within the particle filter algorithm. 
\begin{align}
    \text{bel}(x_{p,t}^k) &= P(x_{p,t}^k|x_{s,t}^k,x_{r,t}^k z_{p,t}^k),\\
    &= \eta P(z_{p,t}^k|x_{p,t}^k, x_{s,t}^k, x_{r,t}^k) \overline{\text{bel}}(x_p(t)), \text{ or}\\
    &= \eta P(\overline{z_{p,t}^k}|x_{p,t}^k, x_{s,t}^k, x_{r,t}^k) \overline{\text{bel}}(x_p(t)).
\end{align}

$P(z_{p,t}^k|x_{p,t}^k, x_{s,t}^k, x_{r,t}^k)$ is used when the object is visible and increases the weights of particles representing poses close to the object's observed pose. Specifically, we employ a Gaussian measurement model of the form:
\begin{align}
    P(z_{p,t}^k | x_{p,t}^{[i], k}, x_{s,t}^k, x_{r,t}^k) = \frac{1}{\eta}\exp (-\frac{d^2}{2\sigma^2}),
\end{align}
where $d$ is the Euclidean distance between the particle's pose and the observed pose. The parameters $\sigma$ and $\eta$ represent the standard deviation of the measurement noise and a normalization constant, respectively.

$P(\overline{z_{p,t}^k}|x_{p,t}^k, x_{s,t}^k, x_{r,t}^k)$ is used when the object is not visible. This function retains the weights of particles that are occluded or outside the visible region, while decreasing the weights of particles that are visible.
\begin{align}
    P(\overline{z_{p,t}^k} | x_{p,t}^{[i], k}, x_{s,t}^k, x_{r,t}^k) = 
    \begin{cases} 
    0, & \text{Visible}, \\
    \frac{1}{\eta}w_{t-1}^{k,[i]}, &\text{Invisible},
    \end{cases}
\end{align}
where $\eta$ is a normalization constant. Similar to the co-location model, when object $k$ is not visible but object $j$ was observed, the weights of the particles are set using the semantic similarity between objects $k$ and $j$ as shown below.
\begin{align}
    w_t^k &= \frac{1 + \texttt{sim}(j,k)}{2} e^{-d/\lambda} + \frac{1-\texttt{sim}(j,k)}{2}(1-e^{-d/\lambda}),
\end{align}
where $d$ is the distance between the detected object and the particle, $\lambda$ is a hyperparameter that determines how fast the weight should increase/decrease with respect to the distance.

\section{Experiments}
\begin{figure}
\centering
    \includegraphics[width=0.8\columnwidth]{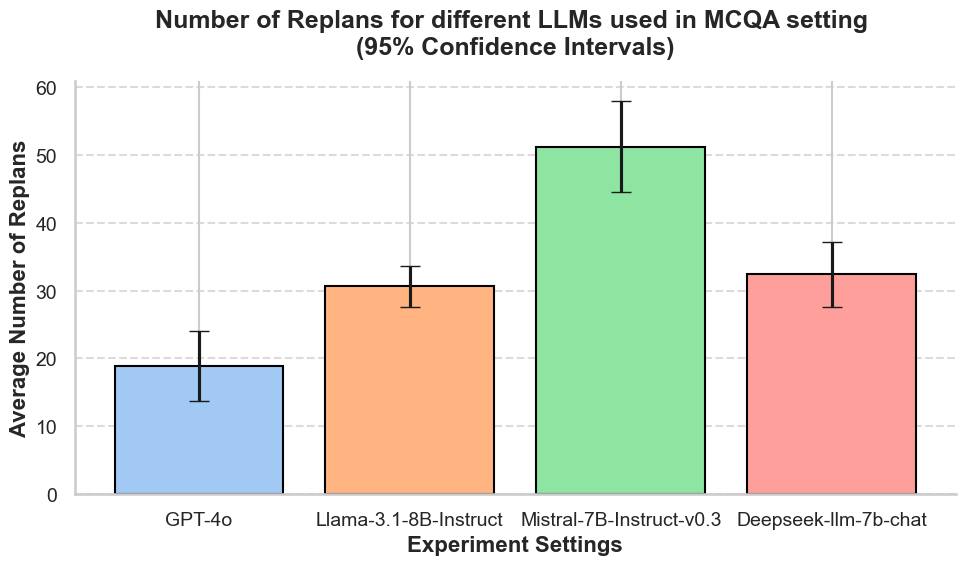} \label{fig:plan_llm}
\caption{Performance metrics over 50 environments in a household layout (6 rooms, 12 surfaces), comparing different LLMs for the MCQA setting. The plots show the number of replans with 95\% confidence intervals. We see GPT-4o outperforms the other smaller models, and hence we have used GPT-4o in all other experiments in this paper.}
\label{fig:llm_performance_combined}
\end{figure}

We evaluate \methodname{} in household environments through both simulation and real-world experiments, measuring cumulative planning and execution time as well as the number of replanning iterations. In simulations, we conduct large-scale experiments leveraging a large household dataset (\ie, the Housekeep dataset~\cite{kant2022housekeep}) to place objects in a common-sense manner, validating the scalable efficiency of \methodname{} compared to baselines. In real-world experiments, we demonstrate \methodname{}’s practical effectiveness by deploying it on Toyota’s human support robot (HSR)~\cite{yamamoto2019development}. In Figure~\ref{fig:llm_performance_combined} we compared initial belief generation across multiple LLM models. As GPT-4o consistently outperformed the alternatives within our evaluation scope, all subsequent experiments were conducted using GPT-4o via API calls.

\subsection{Large-scale simulations}~\label{subsec:sim}

\begin{figure}[t]
  \centering
  \includegraphics[width=0.9\columnwidth]{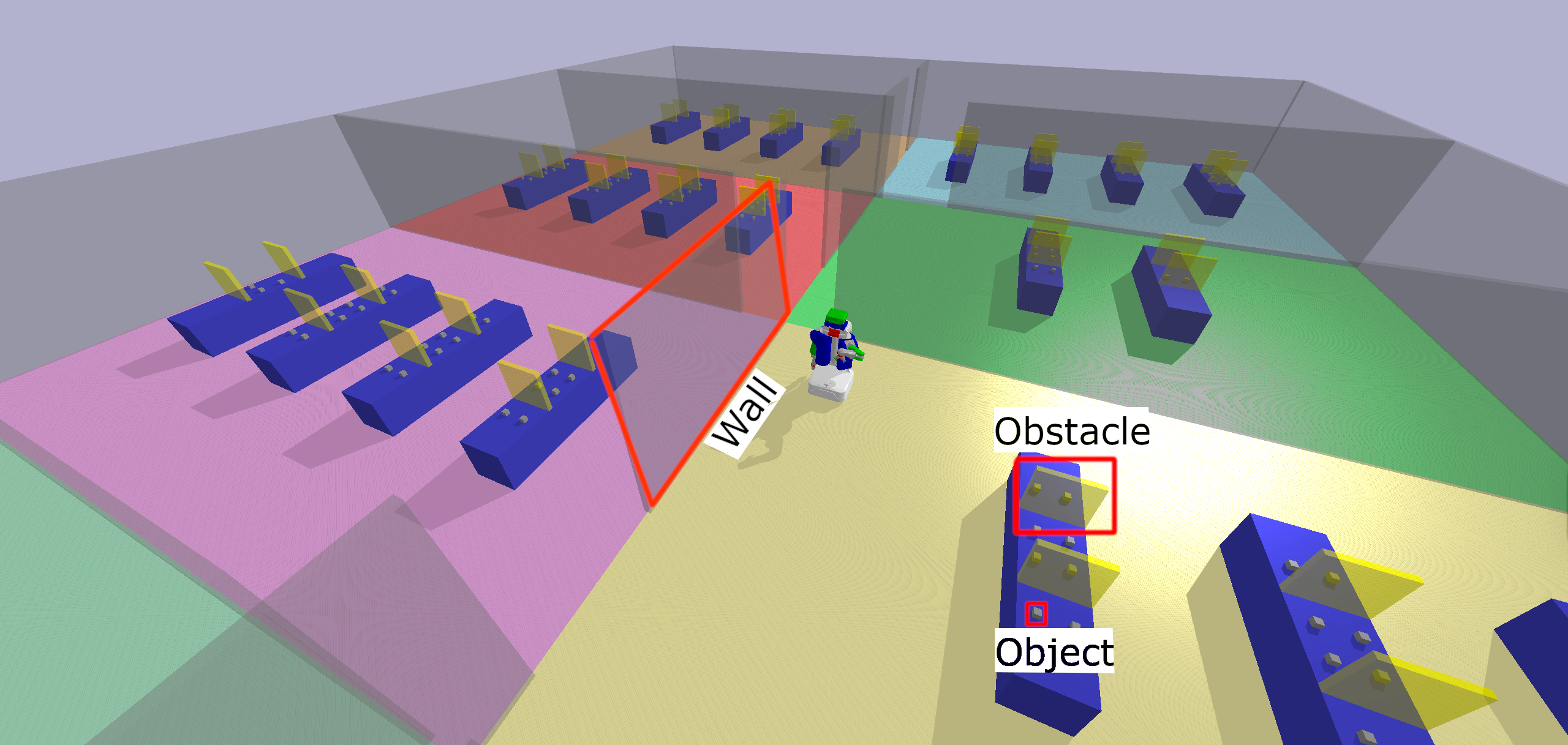}
  \caption{Example of a simulated household environment.}\label{fig:sim}
\end{figure}
\vspace{-0.75\baselineskip}

One challenge in generating simulated household environments is determining where to place objects, as our decisions may introduce bias in favor of our algorithm and fail to reflect common sense. To mitigate this issue, we sample household environments from the existing large-scale Housekeep dataset, which exhibits common-sense object placements. An example of a simulated environment is shown in Fig~\ref{fig:sim}. Note that \methodname{} does not use visual semantics and therefore a realistic simulation is unnecessary. 

\subsubsection{Comparative analysis}\label{subsub:compare}
We evaluate the effectiveness of the proposed LLM-generated initial belief and co-location model by comparing six different variants. Our evaluation metrics include both the cumulative planning and execution time, as well as the number of replans necessary to solve the PO-TAMP problem with \methodname{}. The six methods are: \emph{Baseline}, \emph{Co-Model}, \emph{LLM generated belief update (LGBU)}, \emph{MCQA}, \emph{MCQA with Co-Model}, \emph{MCQA with LGBU}.

These six methods differ in their method for generating the initial belief over the objects' semantic locations  (\ie, $\text{bel}(x_{s,0}^k)$, $\text{bel}(x_{r,0}^k)$) as described in Section~\ref{sub:initial_belief}, and the use of the co-location model as discussed in Section~\ref{subsub:co-model}. In particular, LGBU updates $\text{bel}(x_{s,t}^k)$ and $\text{bel}(x_{r,t}^k)$ directly using LLM-generated predictions based on the current observation using next-token prediction in a manner analogous to initial belief estimation (Section~\ref{sub:initial_belief}). The specific prompts used for LGBU are detailed in Appendix~\ref{appx:lgbu}. Table~\ref{tab:comparison} summarizes these differences by illustrating both the initial belief generation method and the use of the co-location model or LLM-generated update for each method.

\begin{table*}
    \centering 
    \begin{tabular}{|c|c|c|c|c|c|c|}
        \hline
        &\textbf{Baseline} & \textbf{Co-Model} & \textbf{LGBU} &\textbf{MCQA} & \textbf{MCQA with Co-Model} & \textbf{MCQA with LGBU}\\ 
        \hline
        $bel(x_{s,0}^k)$, $bel(x_{s,0}^k)$ & Uniform & Uniform & Uniform & MCQA & MCQA & MCQA\\ 
        \hline
        co-location model & No & Yes & No & No & Yes & No\\
        \hline
        LLM generated update & No & No & Yes & No & No & Yes\\
        \hline
    \end{tabular}
    \caption{Comparison method details.The terms $bel(x_{s,0}^k)$ and $bel(x_{r,0}^k)$ denote the initial belief  over surfaces and rooms, respectively, for object $k$.}\label{tab:comparison}
\end{table*}

\begin{table*}[htbp]
    \centering
    \resizebox{\textwidth}{!}{
    \begin{tabular}{|c|c|c|c|c|c|c|}
        \hline
        & 4 rooms 8 surfs & 4 rooms 16 surfs & 6 rooms 12 surfs & 6 rooms 24 surfs & 8 rooms 16 surfs & 8 rooms 32 surfs \\
        \hline
        \textbf{Baseline} 
        & $632.5 \pm 124.2$ 
        & $715.1 \pm 240.5$ 
        & $1205.3 \pm 272$
        & $1994 \pm 602.7$
        & $1564.9 \pm 362$
        & $3017.5 \pm 1066.1$ \\
        \hline
        \textbf{Co-Model} 
        & $121.2 \pm 58.4$
        & $70.2 \pm 198.5$
        & $185.1 \pm 145.6$
        & $355.3 \pm 398.7$
        & $398.8 \pm 272$
        & $0$ \\
        \hline
        \textbf{MCQA} 
        & $99.4 \pm 95.3$
        & $169.9 \pm 306.4$
        & $248.2 \pm 262.2$
        & $1212.6 \pm 735.1$
        & $134 \pm 231$
        & $1925.3 \pm 1320.8$ \\
        \hline
        \textbf{MCQA with Co-Model} 
        & $0$
        & $0$
        & $0$
        & $0$
        & $0$
        & $50.5 \pm 834.8$ \\
        \hline
        \textbf{LGBU}
        & $164.4 \pm 92.9$
        & $605.8 \pm 288.8$
        & $894.8 \pm 551.9$
        & $509.2 \pm 405$ 
        & $787.2 \pm 321.3$
        & $1066.9 \pm 1054.3$ \\
        \hline
        \textbf{MCQA with LGBU} 
        & $0.2 \pm 48.8$
        & $339.2 \pm 286.5$
        & $101.0 \pm 210.7$
        & $351.1 \pm 361.2$
        & $288.8 \pm 308.3$
        & $408.5 \pm 871.7$\\
        \hline
    \end{tabular}
    }
    \caption{Pairwise comparison of cumulative planning and execution time (in seconds) relative to the best-performing method. Numbers represent the mean $\pm$ 95\% confidence interval calculated over 50 problems, and 0 indicates the best-performing method.}
    \label{tab:CI_time}
\end{table*}

\begin{table*}[htbp]
    \centering
    \resizebox{\textwidth}{!}{
    \begin{tabular}{|c|c|c|c|c|c|c|}
        \hline
        & 4 rooms 8 surfs & 4 rooms 16 surfs & 6 rooms 12 surfs & 6 rooms 24 surfs & 8 rooms 16 surfs & 8 rooms 32 surfs \\
        \hline
        \textbf{Baseline}
        & $9.5 \pm 3.0$
        & $7.1 \pm 6.8$
        & $17.2 \pm 5.4$
        & $25.9 \pm 8.8$
        & $19.8 \pm 6.4$
        & $35.3 \pm 13.7$ \\
        \hline
        \textbf{Co-Model}
        & $3.0 \pm 1.9$
        & $3.4 \pm 5.8$
        & $5.5 \pm 3.1$
        & $7.9 \pm 5.5$
        & $10.6 \pm 6.1$
        & $10.5 \pm 7.4$ \\
        \hline
        \textbf{MCQA}
        & $2.4 \pm 2.7$
        & $0$
        & $2.9 \pm 4.9$
        & $12.8 \pm 8.7$
        & $0$
        & $13.9 \pm 14.1$ \\
        \hline
        \textbf{MCQA with Co-Model}
        & $0.6 \pm 1.4$
        & $0.3 \pm 6.6$
        & $0$
        & $0$
        & $1.4 \pm 4.1$
        & $0$ \\
        \hline
        \textbf{LGBU}
        & $4.1 \pm 2.9$
        & $18.5 \pm 10.7$
        & $23.9 \pm 14.6$
        & $17.9 \pm 10.1$
        & $21.3 \pm 8.3$
        & $15.5 \pm 8.8$ \\
        \hline
        \textbf{MCQA with LGBU}
        & $11.2 \pm 9.4$
        & $0$
        & $3.1 \pm 5.5$
        & $10.2 \pm 7.8$
        & $6.6 \pm 7.0$
        & $6.1 \pm 8.2$\\
        \hline
    \end{tabular}}
    \caption{Pairwise comparison of number of replans relative to the best-performing method. Numbers represent the mean $\pm$ 95\% confidence interval calculated over 50 problems, and 0 indicates the best-performing method.}
    \label{tab:ci_replan}
\end{table*}

\subsubsection{Comparison results}\label{subsub:results}
We design our experiments to test the following hypotheses:
\begin{itemize}
    \item H1: \methodname{} can generalize to diverse common household environments.
    \item H2: Common-sense knowledge about object locations, encoded in LLMs, can improve framework efficiency.
    \item H3: Semantic similarity between objects provides useful inductive bias that can improve framework efficiency.
    \item H4: Solely relying on LLMs for belief updates (LGBU) is insufficient for long-horizon planning and execution.
\end{itemize}

To test Hypothesis 1, we evaluate \methodname{} in a large-scale simulated household environment across six layout configurations varying in the number of rooms and surfaces: {4 rooms and 8 surfaces}, {4 rooms and 16 surfaces}, {6 rooms and 12 surfaces}, {6 rooms and 24 surfaces}, {8 rooms and 16 surfaces}, and {8 rooms and 32 surfaces}. For each layout, we sample 50 distinct environments by varying the labels of rooms, surfaces, and objects while keeping their quantities fixed. Occluding obstacles (\eg, yellow boxes) are added to verify robustness under partial observability.

\begin{figure}[t]
  \centering
  \includegraphics[width=0.8\columnwidth]{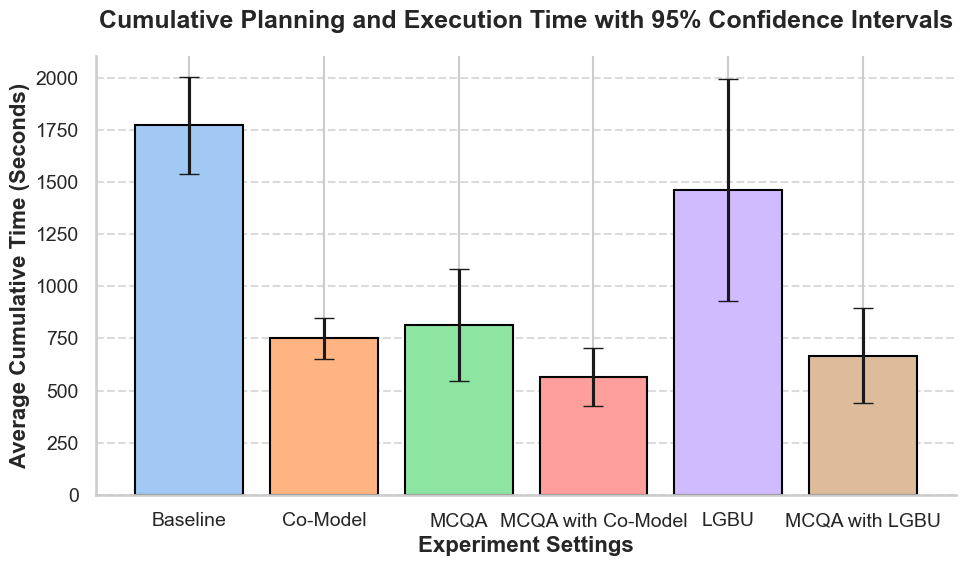}
  \caption{Average cumulative planning and execution time over 50 environments in household layout with 6 rooms and 12 surfaces with 95\% confidence interval.}\label{fig:combined_time}
\end{figure}

\begin{figure}
  \centering
  \includegraphics[width=0.8\columnwidth]{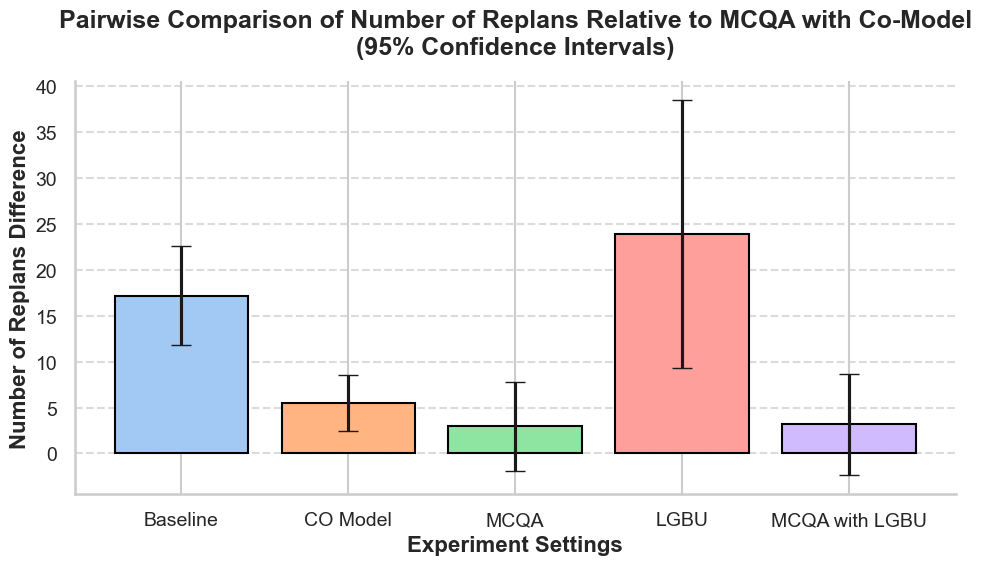}
  \caption{Pairwise comparison of number of replans required to complete the task compared to MCQA with co-location Model with 95\% confidence interval.}\label{fig:additional_replan}
\end{figure}

Given a task, such as detecting and moving an apple to the kitchen table, we record the cumulative planning and execution times and the number of replanning iterations across 50 environments for each layout. We also provide pairwise comparisons with respect to the best-performing method (\ie, the one yielding the lowest cumulative planning and execution times and the fewest replanning iterations). These comparisons indicate, on average, the additional time and extra replanning iterations each method requires relative to the best-performing method. Notably, the best-performing method is deemed statistically superior when the mean value minus its confidence interval remains greater than zero. A comprehensive visualization of the results for layouts consisting of six rooms and twelve surfaces is presented in Figures~\ref{fig:combined_time} and~\ref{fig:additional_replan}, and the full results are provided in Table~\ref{tab:CI_time} and~\ref{tab:ci_replan}.

\begin{figure}
  \centering
  \includegraphics[width=0.8\columnwidth]{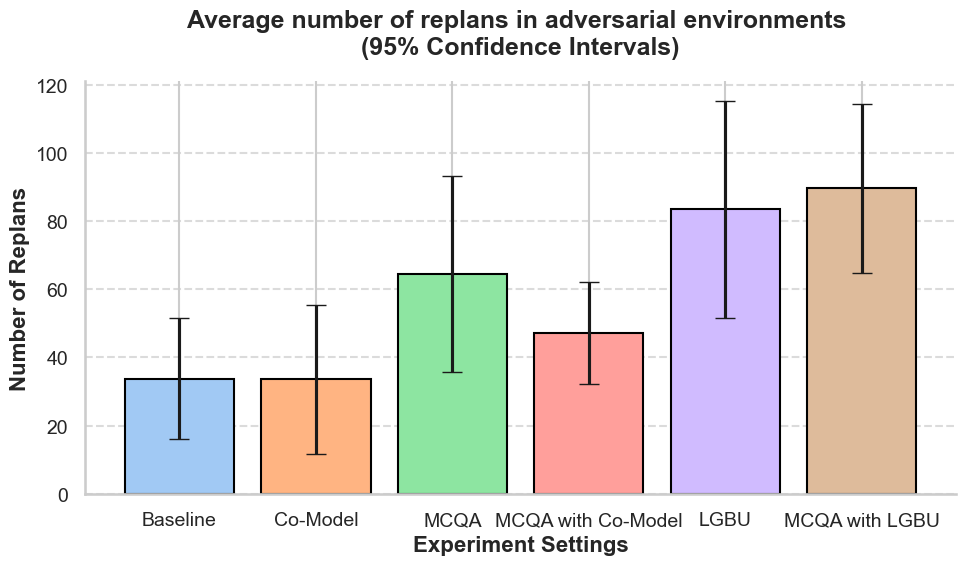}
  \caption{Average number of replans over 5 environments in household layout with 6 rooms and 12 surfaces with 95\% confidence interval.  \\ Note: The LGBU and MCQA with LGBU experiments had a total limit of 100 replans, which was exceeded 3 out of 5 times this experiment was run.}\label{fig:adversarial}
\end{figure}

To test Hypothesis 2, we compare the variant of \methodname{} that uses only an LLM-generated initial belief (via MCQA) to a baseline without such semantic priors. The results show a notable reduction in both cumulative planning and execution time, as well as in the number of replanning iterations, indicating that the LLM prior effectively guides early decision-making. 

To examine Hypothesis 3, a variant that uses only the co-location model during execution is compared against the baseline. This variant reduces the variability (confidence interval) as well as the average of cumulative time and number of replans. This supports that semantic co-locations between objects improve belief refinement throughout long-horizon planning and execution.

Finally, to investigate Hypothesis 4, we evaluate an approach referred to as LGBU, in which belief updates are performed soley using LLM-generated predictions. Experimental results indicate that LGBU results in more frequent replanning and higher cumulative planning and execution time compared to methods that utilize a principled Bayesian belief update, such as the Baseline and Co-Model variants. Figure~\ref{fig:adversarial} further reports results under adversarial configurations, where object placements were randomized to disrupt commonsense regularities. In this setting, Bayesian update methods consistently achieved task completion, whereas LGBU failed in 3 out of 5 trials. These findings indicate that while LLMs provide informative priors, ungrounded LLM-only updates lack the robustness required for reliable long-horizon execution.

Collectively, the results demonstrate that combining the LLM-generated initial belief with the co-location model yields the lowest average number of replanning iterations and cumulative planning and execution time, while also exhibiting the least variability across trials among all evaluated variants. 

\subsection{Real-world experiments}~\label{subsec:real}
We demonstrate \methodname{} in a real-world setting using the HSR robot and a mock apartment environment consisting of two rooms (living room, kitchen) and three surfaces (coffee table, bench, table). 

\begin{figure}[htbp]
  \centering
  \includegraphics[width=0.8\columnwidth]{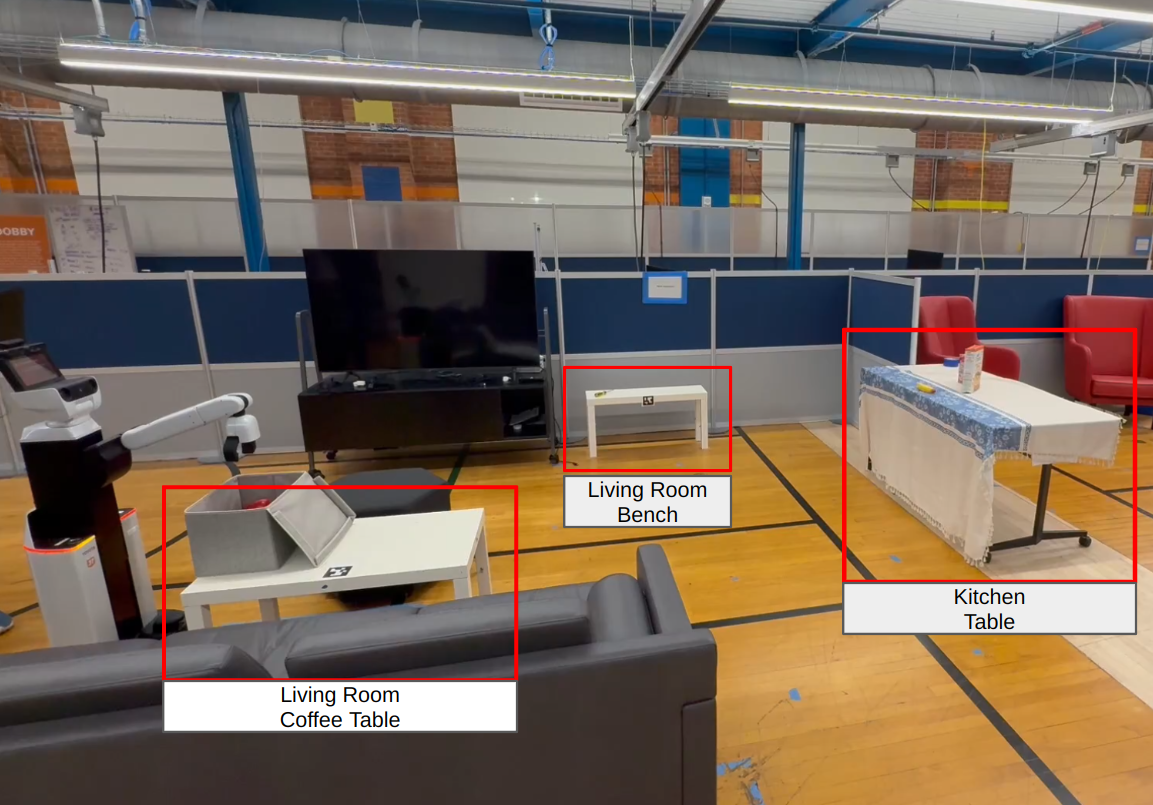}
  \caption{Environment for real-world experiments.}\label{fig:real}
\end{figure}

In this setup, the coffee table and bench are located in the living room, while the table is in the kitchen. The task involves relocating an apple from the kitchen table to the coffee table. Additionally, we introduce a banana and a screwdriver to evaluate the effectiveness of the co-location model and use a cracker box and a cereal box to create occlusions. The real-world setup is shown in Figure~\ref{fig:real}. 

The resulting cumulative planning and execution times are as follows: Baseline: 365 seconds, Co-Model: 112 seconds, and MCQA with Co-Model: 100 seconds.

\section{Related Work}

% TAMP problems are challenging search problems as they combine NP-hard task planning problems~\cite{bylander1992complexity} with PSPACE-hard motion planning problems~\cite{reif1979complexity}.
% Various learning-based approaches have been employed to design efficient TAMP planners, including feasibility prediction at the motion level~\cite{wells2019learning, driess2020deep,li2023sampling,yang2023sequence,sung2023motion}, object importance prediction~\cite{silver2021planning}, guided sampling~\cite{chitnis2016guided,fang2024dimsam}, value function prediction~\cite{kim2022representation}, and learning task-level heuristics~\cite{sung2023learning}. Our work incorporates learning components, informed initial beliefs, and co-locations between objects, achieving these capabilities through few-shot prompting of LLMs that do not require data collection or model training.

Belief space planning aims to handle stochastic transitions and imperfect observations, requiring the solution of POMDP problems. Belief space TAMP planners~\cite{kaelbling2013integrated,fang2024dimsam,curtis2024partially,pan2024task} typically focus on modeling uncertainty at both the task and motion levels while designing approximate solutions to mitigate the complexity of POMDPs. However, they inevitably require replanning to complete tasks within an integrated planning and execution framework. In contrast, \methodname{} aims to enhance the efficiency of this belief space framework by shaping beliefs informed by LLMs.

As our work specifically addresses partial observability and even complete unobservability, the object search literature is also closely related~\cite{zheng2021multi,zheng2022towards,park2023zero,sharma2023semantic,zheng2023asystem}. While these methods tackle certain forms of POMDP problems, they do not fully address TAMP problems, as they do not incorporate task-level symbolic components. Therefore, the object search literature can be considered a subclass of \methodname{}.

With the rise of LLM research, LLMs have been incorporated into TAMP problems in various ways, replacing components that are typically either engineered or learned from data—such as constraints~\cite{curtis2024trust,guo2024castl,kumar2024open}, task-level plan computation~\cite{wang2024llm,li2024league++,ren2023robots,brohan2023can,huang2022language, mohammadi2025more}, or symbolic and geometric
spatial relationships generator ~\cite{ding2023task} —with their common-sense. Additionally, some approaches utilize LLMs to translate natural language descriptions of planning problems into PDDL representations~\cite{liu2023llm+, xie2023translating}. In contrast, \methodname{} leverages LLMs to query a different aspect of commonsense knowledge, specifically related to object placement.

\section{Conclusion} \label{sec:conclusion}

We present \methodname{}, a belief-space planning and execution framework that leverages LLMs to integrate common-sense reasoning into PO-TAMP problems. The framework encodes two key types of common-sense knowledge: (1) certain objects are more likely to be found in specific locations, and (2) semantically similar objects tend to be co-located, while dissimilar ones are not. While LLMs likely possess transferable knowledge applicable to non-household domains (\eg, factories or hospitals), \methodname{} has not yet been evaluated in such settings. 
% Additionally, \methodname{} performance degrades in adversarial environments where common-sense does not hold.
In some scenarios, full information about the environment layout may be unavailable, necessitating information gathering about rooms and surfaces. We leave these directions for future work, which we are currently pursuing.

%%%%%%%%%%%%%%%%%%%%%%%%%%%%%%%%%%%%%%%%%%%%%%%%%%%%%%%%%%%%%%%%%%%%%%%%%%%%%%%%

%%%%%%%%%%%%%%%%%%%%%%%%%%%%%%%%%%%%%%%%%%%%%%%%%%%%%%%%%%%%%%%%%%%%%%%%%%%%%%%%

%%%%%%%%%%%%%%%%%%%%%%%%%%%%%%%%%%%%%%%%%%%%%%%%%%%%%%%%%%%%%%%%%%%%%%%%%%%%%%%%

\section*{ACKNOWLEDGMENT}

This work has taken place in the Learning Agents Research
Group (LARG) at the Artificial Intelligence Laboratory, The University of Texas at Austin.  LARG research is supported in part by the National Science Foundation (FAIN-2019844, NRT-2125858), the Office of Naval Research (N00014-24-1-2550), Army Research Office (W911NF-17-2-0181, W911NF-23-2-0004, W911NF-25-1-0065), DARPA (Cooperative Agreement HR00112520004 on Ad Hoc Teamwork), Lockheed Martin, and Good Systems, a research grand challenge at the University of Texas at Austin.  The views and conclusions contained in this document are those of the authors alone.  Peter Stone serves as the Chief Scientist of Sony AI and receives financial compensation for that role.  The terms of this arrangement have been reviewed and approved by the University of Texas at Austin in accordance with its policy on objectivity in research.

%%%%%%%%%%%%%%%%%%%%%%%%%%%%%%%%%%%%%%%%%%%%%%%%%%%%%%%%%%%%%%%%%%%%%%%%%%%%%%%%

\bibliographystyle{IEEEtran}
\bibliography{icra2026}
% \addtolength{\textheight}{-12cm} 
% This command serves to balance the column lengths on the last page of the document manually. It shortens the textheight of the last page by a suitable amount. This command does not take effect until the next page so it should come on the page before the last. Make sure that you do not shorten the textheight too much.

\onecolumn
\section*{APPENDIX}

\subsection{Co-location Model Prompt}\label{appx:co_model_prompt}
\textbf{Prompt:} Explain the common use of a toaster in three sentences.\\
\textbf{LLM Response} (GPT4o): A toaster is commonly used to toast slices of bread, making them crispy and golden brown. It's also useful for heating up and adding a slight crunch to bagels, English muffins, and some pastries. Additionally, many people use a toaster to quickly prepare breakfast items like Pop-Tarts or frozen waffles.

\subsection{Co-Location Toggler Prompt}\label{appx:toggle_prompt}
\textbf{System Prompt}: You will be given an object that commonly appears in a typical household environment. Using common sense, determine if the object tends to be distributed throughout a typical household, such as doorknobs and light switches.\\
\textbf{User Prompt}: The following is the object: [object]\\
\textbf{LLM Response} (GPT4o): True / False

\subsection{LGBU Prompt}\label{appx:lgbu}
\textbf{System Prompt}: You will receive:
\begin{itemize}
    \item Current belief about where an object might be located.
    \item observation\_location: the location that was just inspected.
    \item visibility: how much of the location was visible. 0 means not visible at all, 1 means fully visible.
    \item result: whether the object was found there.
    \item co\_detected: other objects found at the same location.
\end{itemize}
Based on this information and common sense, predict where the object is most likely to be now. Choose the most likely location from the given options.\\
\textbf{User Prompt}: 
\begin{itemize}
    \item Current belief about [object\_name]: [belief\_text]
    \item observation\_location: [obs\_loc]
    \item visibility: [visibility]
    \item result: [result]
    \item co\_detected: [co\_detected\_text]
\end{itemize}
Given this information, where is {object\_name} most likely to be? [options\_text]\\
Similar to the MCQA method, the belief is updated using next token prediction, where [option\_text] is a multiple choice with a character paired with a surface or room.

\subsection{Detect action}\label{appx:detect}
To address partial observability, we introduce the \texttt{detect} action as part of the action set $\mathcal{A}$, alongside conventional actions commonly used in PDDLStream, such as \texttt{pick}, \texttt{place}, \texttt{move}, and their corresponding streams.

The \texttt{detect} action is parameterized by the object \texttt{?o}, surface \texttt{?s}, room \texttt{?r}, sampled object pose from $\text{bel}(x_{p,t}^k)$ \texttt{?pb}, base configuration \texttt{?bq}, head configuration \texttt{?hq}, and head trajectory \texttt{?ht} of the robot. 

\begin{lstlisting}[language=PDDL]
(:action detect
 :param (?o ?s ?r ?pb ?bq ?hq ?ht)
 :pre (and (IsItem ?o) (IsSurf ?s) (IsRoom ?r)
           (AtBConf ?bq) (PoseB ?o ?pb ?s)
           (Vis ?o ?pb ?bq ?hq ?ht))
 :eff (and (Supported ?o ?pb ?s) (AtPoseB ?o ?pb)
           (increase (total-cost) 
           (DetectCost ?o ?s ?r ?pb))))
\end{lstlisting}

Among the preconditions listed above, \texttt{(PoseB ?o ?pb ?s)} and \texttt{(Vis ?o ?pb ?pb ?hq ?ht)} are predicates that must be verified by streams due to the presence of continuous typed variables. 

The \texttt{sample-PoseB} stream samples a pose \texttt{?pb} for an object \texttt{?o} from a specified surface \texttt{?s}. The stream then certifies that \texttt{?pb} is a valid sampled pose of object \texttt{?o} on surface \texttt{?s} by generating the predicate \texttt{(PoseB ?o ?pb ?s)}, which is one of the preconditions for the \texttt{detect} action.
\begin{lstlisting}[language=PDDL]
(:stream sample-PoseB
 :inp (?o ?s)
 :dom (and (IsItem ?o) (IsSurf ?s))
 :out (?pb)
 :cert (and (PoseB ?o ?pb ?s)))
\end{lstlisting}

The \texttt{inverse-visibility} stream produces a robot configuration ensuring that the sampled object pose \texttt{?pb} remains within the robot's field of view. Specifically, \texttt{?pb} is considered visible if it lies within the robot’s predefined visibility cone. This stream guarantees that the robot’s orientation and proximity are appropriate for observing the location where the object is believed to be, by generating the predicate \texttt{(Vis ?o ?pb ?bq ?hq ?ht)}.
\begin{lstlisting}[language=PDDL]
(:stream inverse-visibility
:inp (?o ?pb ?s)
:dom (and (IsItem ?o) (IsSurf ?s) (PoseB ?o ?pb ?s))
:out (?hq ?ht ?bq)
:cert (and (Vis ?o ?pb ?bq ?hq ?ht)))
\end{lstlisting}

The preconditions \texttt{(IsItem ?o)}, \texttt{(IsSurf ?s)}, and \texttt{(IsRoom ?r)} state that the parameters \texttt{?o}, \texttt{?s}, and \texttt{?r} correspond to a specific object, surface, and room, respectively. \texttt{(AtBConf ?bq)} asserts that the robot’s base configuration is \texttt{?bq}.

The effects of the \texttt{detect} action are defined as follows: \texttt{(AtPoseB ?o ?pb)} asserts that the object \texttt{?o} is at the sampled pose \texttt{?pb}, and \texttt{(Supported ?o ?pb ?s)} asserts that the sampled pose \texttt{?pb} of object \texttt{?o} is on the surface \texttt{?s}.

\subsection{Object State Estimation Derivation}\label{appx:obj_sed}
\begin{align}
    &\text{bel}(x_{r,t}^k, x_{s,t}^k, x_{p,t}^k) = P(x_{r,t}^k, x_{s,t}^k, x_{p,t}^k|\mathcal{Z}_{r,1:t},\mathcal{Z}_{s,1:t}, z_{p,1:t}^k, v_{r,1:t}, v_{s,1:t})\\
    &= P(x_{p,t}^k|x_{r,t}^k, x_{r,t}^k, \mathcal{Z}_{r,1:t},\mathcal{Z}_{s,1:t}, z_{p,1:t}^k, v_{r,1:t}, v_{s,1:t})P(x_{r,t}^k, x_{s,t}^k|\mathcal{Z}_{r,1:t},\mathcal{Z}_{s,1:t},z_{p,1:t}^k,v_{r,1:t}, v_{s,1:t})\\
    &= P(x_{p,t}^k|x_{r,t}^k, x_{r,t}^k, z_{p,1:t}^k)P(x_{r,t}^k, x_{s,t}^k|\mathcal{Z}_{r,1:t},\mathcal{Z}_{s,1:t},v_{r,1:t}, v_{s,1:t})\\
    &=P(x_{p,t}^k|x_{r,t}^k, x_{r,t}^k, z_{p,1:t}^k)P(x_{s,t}^k| x_{r,t}^k,\mathcal{Z}_{r,1:t},\mathcal{Z}_{s,1:t},v_{r,1:t}, v_{s,1:t})P(x_{r,t}^k|\mathcal{Z}_{r,1:t},\mathcal{Z}_{s,1:t},v_{r,1:t}, v_{s,1:t})\\
    &=P(x_{p,t}^k|x_{r,t}^k, x_{r,t}^k, z_{p,1:t}^k) P(x_{s,t}^k|x_{r,t}^k, \mathcal{Z}_{s,1:t}, v_{s,1:t}) P(x_{r,t}^k|\mathcal{Z}_{r,1:t}, v_{r,1:t})\\
    &=P(x_{p,t}^k|x_{r,t}^k, x_{r,t}^k, z_{p,t}^k) P(x_{s,t}^k|x_{r,t}^k, \mathcal{Z}_{s,t}, v_{s,t}) P(x_{r,t}^k|\mathcal{Z}_{r,t}, v_{r,t})
\end{align}
\begin{itemize}
    \item (1) - (2): Chain rule.
    \item (2) - (3): Conditional independence assumption. 
Once the room and surface where the object is located($x_{r,t}^k, x_{s,t}^k$) are known, further observations over rooms and surfaces($\mathcal{Z}_{r,1:t},\mathcal{Z}_{s,1:t}$) do not contribute additional information. Similarly, the visibility of the room and the surface ($v_{r,1:t} s_{r,1:t}$) does not contribute to estimating the pose of the object of interest.
    \item (3) - (4): Chain rule.
    \item (4) - (5): Conditional independence assumption. Once which room the object is located ($x_{r,t}^k$) is known, further observations over rooms ($\mathcal{Z}_{r,1:t}$) and the visibility of that room ($v_{r,t}$) does not provide additional information. Furthermore, the estimation of the object's room location depends solely on room observations($\mathcal{Z}_{r,1:t}$) and the room's visibility($v_{r,t}^k$).
    \item (5) - (6): Markov assumption.
\end{itemize}

\subsection{Belief over Surfaces}\label{appx:surf_eqations}

\begin{align}
    \text{bel}(x_{s,t}^k) &= P(x_{s,t}^k|x_{r,t}^k,\mathcal{Z}_{s, t}, v_{s,t}),\\
    &= \eta P(\mathcal{Z}_{s,t}|x_{s,t}^k, x_{r,t}^k, v_{s,t}) \overline{\text{bel}}(x_{s,t}^k).
\end{align}

\begin{equation}
\begin{aligned}
P(\mathcal{Z}_{s,t}\mid x_{s,t}^k,x_{r,t}^k,v_{s,t})
  &= P(z_{s,t}^k \mid x_{s,t}^k,x_{r,t}^k,v_{s,t}) \times
     \prod_{\substack{j=1\\ j\neq k}}^{K}
     P\!\bigl(z_{s,t}^j\mid x_{s,t}^k,x_{r,t}^k\bigr).
\end{aligned}
\end{equation}

\textbf{Case 1:} Observation of the surface where the object is placed (i.e., $z_{s,t}^k = x_{s,t}^k$).
\begin{align}
    P(z_{s,t}^k|x_{s,t}^k, x_{r,t}^k, v_{s,t})=
    \begin{cases}
    (1 - p_{\text{fn}})\, v_{s,t}, & \text{Detected},\\
    (1 - v_{s,t}) + v_{s,t}\, p_{\text{fn}}, & \text{Not detected}.
    \end{cases}
\end{align}

\textbf{Case 2:} Observation of a surface where the object is not placed (i.e., $z_{s,t}^k \neq x_{s,t}^k$).
\begin{align}
    P(z_{s,t}^k|x_{s,t}^k, x_{r,t}^k, v_{s,t})=
    \begin{cases}
    p_{\text{fp}}\, v_{s,t}, & \text{Detected},\\
    1 - v_{s,t} p_{\text{fp}}, & \text{Not detected}.
    \end{cases}
\end{align}

\begin{equation}
\begin{aligned}
    P(z_{s,t}^j|x_{s,t}^k, x_{r,t}^k) = \sum_{\substack{x_{r,t}^j \in \mathcal{R} \\ x_{s,t}^j \in \mathcal{S}}} P(z_{s,t}^j|x_{s,t}^j, x_{r,t}^j) \times P(x_{s,t}^j,x_{r,t}^j | x_{s,t}^k,x_{r,t}^k).
\end{aligned}
\end{equation}

The definition of $P(x_{s,t}^j,x_{r,t}^j | x_{s,t}^k,x_{r,t}^k)$ is analogous.

\subsection{Occlusion Aware Particle Filter Algorithm}\label{appx:particle}
\begin{algorithm}\label{algo:particle_filter}
\caption{Particle filter for object pose estimation}
\begin{algorithmic}[1]
\Procedure{Particle Filter}{$X_{p,t-1}^k, z_{p,t}^k, \overline{z_{p,t}^k}$}
    \State $X_{p,t}^k \gets \emptyset, \overline{X}_{p,t}^k \gets \emptyset$
    \For{$i = 1$ to $N$}
        \State $x_{p,t}^{k,[i]} \gets x_{p,t-1}^{k,[i]}$ \Comment{Stationary objects}
        \If{object $k$ is detected}
            \State $w_t^{k,[i]} \gets P(z_{p,t}^k|x_{p,t}^{k,[i]}, x_{s,t}^k, x_{r,t}^k)$ \Comment{Eq.~13}
        \ElsIf{object $j \neq k$ is detected}
            \State $w_t^{k,[i]} \gets P(\overline{z_{p,t}^k}|x_{p,t}^{k,[i]}, x_{s,t}^k, x_{r,t}^k) \cdot w_t^{k,[i]}$ \Comment{Eq.~14,~15}
        \Else
            \State $w_t^{k,[i]} \gets P(\overline{z_{p,t}^k}|x_{p,t}^{k,[i]}, x_{s,t}^k, x_{r,t}^k)$ \Comment{Eq.~14}
        \EndIf
        \State $\overline{X}_{p,t}^k \gets \overline{X}_{p,t}^k \cup \langle x_{p,t}^{k,[i]}, w_t^{k,[i]}\rangle$
    \EndFor
    \For{$i = 1$ to $N$}
        \State draw $j$ with probability $\propto w_t^{k,[j]}$
        \State $X_{p,t}^k \gets X_{p,t}^k \cup x_{p,t}^{k,[j]}$
    \EndFor
\EndProcedure
\end{algorithmic}
\end{algorithm}

\subsection{Simulation Environment Generation}\label{appx:sim_gen}

The dataset includes 1,799 object categories, 585 surface types, and 15 room types. For each object in a room, 10 human annotators rank the surfaces where a human would most likely place the item, resulting in 45,556 entries. An example of such a data point is as follows: \\
\texttt{\textbf{Object}: multiport hub | \textbf{Room}: kitchen\\
\textbf{Correct} (ranked): carpet, fridge, table, counter, sink\\ 
\textbf{Incorrect} (ranked): chest, cooktop, microwave, dishwasher, stove\\
\textbf{Implausible}: oven, shelf, top cabinet, chair, bottom cabinet}

The \textit{correct} label designates surfaces deemed preferable, the \textit{incorrect} label identifies surfaces where an object is unlikely to be placed, and the \textit{implausible} label refers to surfaces where object placement is considered nearly impossible by humans. Surfaces with the \textit{correct} label are ranked from most to least likely, while those with the \textit{incorrect} label are ranked from least to most likely.

Since the 10 annotators may have different opinions about object placement (\eg, the most likely to least likely surface placements of a \texttt{multiport hub} in a \texttt{kitchen}), we combine the annotations by assigning scores to surfaces in the correct and incorrect labels and summing the scores from all 10 annotators. If more than half of the annotators considered a surface implausible, we removed it from the dataset.

For a data point comprising $N$ surfaces, we define a scoring scheme where the maximum attainable score is $N$ and the minimum attainable score is $-N$. Surfaces labeled as \textit{correct} begin with a score of 
$N$ for the most likely surface, decreasing by 1 for each subsequently less likely surface. In contrast, surfaces labeled as \textit{incorrect} start with a score of $-N$ for the least likely surface, increasing by 1 for each surface deemed more likely. The following example demonstrates this assignment when $N=15$:\\
\texttt{\textbf{Object}: multiport hub | \textbf{Room}: kitchen\\
\textbf{Correct}: carpet=15, fridge=14, table=13, counter=12, sink=11\\
\textbf{Incorrect}: chest=-15, cooktop=-14, microwave=-13, dishwasher=-12, stove=-11}

To integrate the perspectives of the 10 annotators, we aggregate the scores to produce a cumulative value for each object placement. We then organize the data according to a \emph{Room}~\(\rightarrow\) \emph{Surface}~\(\rightarrow\) \emph{Object} hierarchy. Finally, we normalize these aggregated scores to the interval \([0,1]\), yielding a probabilistic representation of the likelihood that an object will be placed on a specific surface within a particular room.

To construct a ground-truth state of a typical household environment, we randomly sample a predefined number of rooms and the surfaces within each room. We then sample a predefined number of objects for each surface, using the probabilities derived from the aforementioned scoring scheme. An example of a sampled household environment in simulation is shown in Figure~\ref{fig:sim}, where the environment is partitioned into rooms by walls, blue boxes denote surfaces, yellow boxes indicate obstacles, and gray boxes represent objects. 

\section{Real World Experiment Analysis}\label{appx:real_results}

Both the Baseline and Co-Model methods first invoke the \texttt{detect} action on the coffee table, as they begin with a uniform prior distribution across surfaces and rooms. Conversely, MCQA with Co-Model initiates \texttt{detect} on the table, informed by the LLM's common sense that the apple is likely to be on the kitchen table. While the Baseline method cannot derive information from other objects, both Co-Model and MCQA with Co-Model detect a similar object (a banana) and adjust the beliefs accordingly, prompting the robot to \texttt{detect} the table from a different viewpoint.

We report the initial and goal literals used in the real-world experiment, enumerate the \texttt{detect} actions executed by each method, and explain the reasons for any \texttt{detect} failures.
\begin{lstlisting}[language=PDDL]
Initial literal:
(IsItem ?apple),(IsItem ?banana),(IsItem ?cracker_box),
(IsItem ?cereal_box),(IsItem ?screwdriver),
(IsSurf ?coffee_table),(IsSurf ?bench),(IsSurf ?table), 
(IsRoom ?living_room),(IsRoom ?kitchen),
(HandEmpty ?arm), (AtBConf ?start)
Goal literal:
(At ?apple ?coffee_table)
\end{lstlisting}

\begin{lstlisting}[language=PDDL]
Baseline:
detect ?apple ?coffee_table ?living_room (1)
detect ?apple ?table ?kitchen            (2)
detect ?apple ?bench ?living_room        (3)
detect ?apple ?coffee_table ?living_room (4)
detect ?apple ?table ?kitchen            (5)
Num of Replan: 4
\end{lstlisting}
\begin{itemize}
    \item (1): Failed due to the apple not being placed on the coffee table.
    \item (2): Failed due to occlusion caused by the cracker and cereal boxes.
    \item (3): Failed due to the apple not being placed on the bench.
    \item (4): Failed due to apple not being placed on the coffee table.
    \item (5): Successfully detected the apple.
\end{itemize}
\begin{lstlisting}[language=PDDL]
Co-Model:
detect ?apple ?coffee_table ?living_room (1)
detect ?apple ?table ?kitchen            (2)
detect ?apple ?table ?kitchen            (3)
Num of Replan: 2
\end{lstlisting}
\begin{itemize}
    \item (1): Failed due to the apple not being placed on the coffee table.
    \item (2): Failed due to occlusion caused by the cracker and cereal boxes. A banana, similar to an apple, was detected instead, and the correlation model updated the belief accordingly.
    \item (3): Successfully detected the apple.
\end{itemize}
\begin{lstlisting}[language=PDDL]
MCQA with Co-Model:
detect ?apple ?table ?kitchen (1)
detect ?apple ?table ?kitchen (2)
Num of Replan: 1
\end{lstlisting}
\begin{itemize}
    \item (1):Failed due to occlusion caused by the cracker and cereal boxes. A banana, similar to an apple, was detected instead, and the correlation model updated the belief accordingly.
    \item (2): Successfully detected the apple.
\end{itemize}

\end{document}